\documentclass[letterpaper, 10 pt, conference]{ieeeconf}  

\IEEEoverridecommandlockouts                              

\usepackage{graphics} 
\usepackage{epsfig} 
\usepackage{amsmath} 
\usepackage{amssymb}  
\usepackage{booktabs}
\usepackage{multirow}
\usepackage{subcaption}
\usepackage{balance}
\usepackage{url}
\usepackage{float}
\usepackage[ruled,linesnumbered,vlined]{algorithm2e}
\usepackage{xcolor}

\usepackage[table]{xcolor} 
\usepackage{nicematrix}

\usepackage{hyperref}
\hypersetup{colorlinks,urlcolor=red}

\usepackage[backend=bibtex,natbib=true,style=numeric-comp,sorting=none,giveninits=true,maxbibnames=1,minbibnames=1,url=false,doi=false]{biblatex}
\SetKwInput{KwIn}{Input}
\SetKwInput{KwOut}{Output}
\SetArgSty{textnormal} 

\title{\LARGE \bf
SAGE-Nav: Leveraging LLM Planning and Alignment Fusion for Hierarchical Scene Graph-Guided Navigation
}

\author{Hao Su$^{1*}$, Yuehao Huang$^{1*}$,  Yukai Ma$^{1}$, Yong Liu$^{1\dagger}$ and Jiajun Lv$^{1\dagger}$
\thanks{$^{1}$All authors are with Zhejiang University, Hangzhou, China. }
\thanks{$^{*}$Equal contribution.}
\thanks{$^\dagger$Corresponding authors. Email: {lvjiajun314@zju.edu.cn}.} 
\thanks{This work was supported by the Key R\&D Project of Zhejiang Province under Grant 2025C01090.}%
}

\begin{document}
\maketitle
\thispagestyle{empty}
\pagestyle{empty}
\begin{refsection} 
\begin{abstract}

Object-Goal Navigation (ObjNav) requires embodied agents to autonomously locate specified targets using only egocentric visual observations. Existing monolithic methods struggle with long-horizon reasoning and generalize poorly to novel environments. To address these limitations, we propose SAGE-Nav, a novel hierarchical framework that integrates the reasoning capabilities of Large Language Models (LLMs) with dynamic scene graphs. Crucially, it decouples asynchronous global semantic planning from the high-frequency reactive control loop. The LLM serves as a global planner, decomposing abstract instructions into a sequence of semantically grounded waypoints. To translate these plans into dense multi-modal guidance, we design a Hierarchical Scene Graph Encoder (HSGE) that leverages relational graph convolutions to produce structure-aware embeddings preserving both semantic and spatial topology. Furthermore, we develop the Goal-aware Alignment-Fusion Network (GAFN) to dynamically fuse real-time perception with these structural priors. Using an adaptive gating mechanism with an explicit inductive bias, GAFN ensures robust visual-topological alignment for the low-level policy. Extensive evaluations in the i-THOR and RoboTHOR environments demonstrate that SAGE-Nav achieves state-of-the-art performance, delivering substantial gains in navigation efficiency and zero-shot generalization while maintaining the low control latency required for physical robotic deployment.

\end{abstract}

\section{INTRODUCTION}

Object-Goal Navigation (ObjNav)~\cite{objectnavsurvey2024,batra2020objectnav,chaplot2020object} 
requires an embodied agent to autonomously locate and reach a specified object based solely on egocentric visual observations, making it a fundamental yet challenging task in embodied AI. 
Deep Reinforcement Learning (DRL) has empowered agents to learn end-to-end navigation policies~\cite{targetdriven2017,SP} directly from egocentric visual inputs, achieving efficient exploration and object-goal approach with asynchronous frameworks~\cite{A3C}. 
However, the navigation policies fundamentally optimize reactive mappings conditioned on immediate visual contexts, without explicit modeling of scene-level semantics. Consequently, when deploying in unseen environments, performance degrades due to distribution shifts, which hinders the target plausible location inference and out-of-view long-horizon reasoning. This challenge has also motivated recent efforts toward unified embodied navigation paradigms~\cite{GN0}, which emphasize the importance of data generation, simulation, evaluation, and policy learning for long-horizon generalization.
Drawing inspiration from human spatial cognition, where navigation relies on hierarchical planning through semantic  scene understanding and abstract spatial reasoning, 
recent works~\cite{HOZ,AKGVP,ContextAwareGI,TemporalSG} introduce structured graph representations to capture semantic and spatial relationships among objects. 
Despite enhancing scene understanding and goal inference, these graph-based methods perform \emph{implicit} end-to-end planning, lacking explicit hierarchical guidance. 
Moreover, the knowledge embedded in learned object co-occurrence maps represents aggregated scene statistics rather than structured, query-dependent planning signals. 
This forces low-level DRL policies to infer long-horizon strategies directly from fused features, compromising interpretability and leaving the agent without progress feedback during execution.

The core problem lies in the absence of high-level semantic reasoning, which is essential for robust target localization in novel scenes. Large Language Models (LLMs) offer this capability through the vast commonsense priors, yet they lack spatial grounding required for navigation. 
To bridge this gap, we introduce a hierarchical framework that integrates LLM-driven planning with the spatially grounded structure of Scene Graphs (SGs).

Prior approaches either (i) leverage LLMs to generate low-level action sequences with imitation learning~\cite{IL_2019_CVPR,IL_2020_RAL,IL_2023_CVPR}, or (ii) apply LLMs directly for zero-shot planning~\cite{OptimalSG,SG-Nav,UniGoal}.
In contrast, our method constructs a hierarchical scene graph as an explicit environment prior. 
The LLM then utilizes Retrieval-Augmented Generation (RAG)~\cite{RAG,leapad,leapvad} to reason with this structured representation, producing an interpretable sequence of semantically grounded waypoints that are spatially constrained, semantically verifiable, and robust.
Crucially, we distill LLM-derived semantic knowledge into the low-level policy rather than repeatedly invoking LLMs at runtime. This allows the system to benefit from strong semantic priors regarding affordances and functional relations, while maintaining the efficiency and responsiveness characteristic of DRL-based agents. 
To ensure global semantic plans are effectively incorporated into the perception level, we introduce the Goal-aware Alignment-Fusion Network (GAFN), which dynamically modulates the fusion between egocentric visual features and scene-graph representations based on waypoint alignment, yielding a continuous and interpretable progress signal for long-horizon execution.
Inspired by recent progress in LLM-enhanced scene graph reasoning~\cite{LLMenhancedSG}, 
we further develop a Hierarchical Scene Graph Encoder (HSGE) that grounds the planned waypoints in the graph structure and produces goal-conditioned representations that preserve both semantic abstraction and spatial topology.

In summary, the contributions of this work are threefold:
\begin{itemize}
    \item We propose SAGE-Nav, a hierarchical navigation paradigm integrating LLM-driven planning with dynamic scene graphs. It decomposes abstract instructions into semantic waypoints, effectively decoupling asynchronous global reasoning from high-frequency control.
    \item We design the Hierarchical Scene Graph Encoder (HSGE) to translate abstract plans into actionable topological guidance. By leveraging relational graph convolutions, it produces structure-aware embeddings designed to capture both semantic and spatial hierarchies.
    \item We develop the Goal-aware Alignment-Fusion Network (GAFN) to dynamically fuse real-time perception with structural priors. Its adaptive gating mechanism introduces an explicit inductive bias, promoting robust visual-topological alignment for policy execution.
\end{itemize}

\section{RELATED WORKS}

\subsection{Object-Goal Navigation}
Object-Goal Navigation requires an agent to locate a specified target using egocentric observations. While deep reinforcement learning has enabled direct mappings from vision to navigation actions~\cite{targetdriven2017,ORG,SAVN,VTNet}, these end-to-end policies often struggle with long-horizon exploration due to limited spatial and semantic priors. To mitigate this, hierarchical and graph-based approaches~\cite{HOZ,SP,DOA,AKGVP} have been proposed to capture object relations, regional connectivity, and graph-visual alignment. Recent LLM-guided methods~\cite{SayNav,CogNav, CogDDN, UcON, NavFoM, Navid, Uni-navid, AURA, DeCoNav} further introduce commonsense reasoning for semantic object search, but often rely on frequent foundation-model queries during execution. In contrast, SAGE-Nav decouples global semantic reasoning from local reactive control, leveraging an LLM-driven planner to generate semantically grounded waypoints that are encoded and aligned with real-time perception for robust execution.

\subsection{Scene Graph Representations in Navigation}
Vision-Language Models~\cite{CLIP,llama} have established Scene Graphs as a paradigm for endowing agents with structured world knowledge~\cite{conceptgraphs}. Recent works~\cite{OptimalSG,SG-Nav,HovSG} integrate LLMs with scene graphs for reasoning, where SG-Nav~\cite{SG-Nav} prompts an online 3D scene graph for zero-shot object navigation. Beyond navigation, language-grounded hierarchical planning with multi-robot 3D scene graphs~\cite{LGHP} demonstrates the value of shared scene graphs for language-conditioned task execution. Building upon these efforts, SAGE-Nav uses hierarchical scene graphs not only as an LLM planning substrate but also as structured priors encoded by HSGE and fused with egocentric perception through GAFN. By distilling abstract reasoning into grounded waypoints, SAGE-Nav provides the downstream reactive policy with interpretable structural guidance.

\section{TASK DEFINITION}

We formalize Object-Goal Navigation as a Partially Observable Markov Decision Process (POMDP), defined by $\langle \mathcal{S}, \mathcal{A}, \mathcal{T}, \mathcal{R}, \Omega, \mathcal{O}, \gamma \rangle$. Given a task instruction $q \in \mathcal{Q}$ specifying the target object, the agent seeks a policy $\pi(a_t \mid h_t, q)$ to maximize the expected cumulative reward. At each step $t$, the agent at state $s_t = (x, z, \theta_{\text{yaw}}, \theta_{\text{pitch}}) \in \mathcal{S}$ receives an egocentric observation $o_t \in \Omega$ via $\mathcal{O}(o_t \mid s_t)$. 
The agent then selects an action $a_t$ from the discrete space $\mathcal{A} = \{\text{MoveAhead}, \text{RotateL/R}, \text{LookU/D}, \text{Done}\}$.
The environment evolves according to transition dynamics $\mathcal{T}(s_{t+1} \mid s_t, a_t)$. 
An episode is considered successful when the agent selects the action \texttt{Done} within $T_{\max}$ steps, provided that the target $p$ is visible and within a distance $d_s$ (e.g., $1.5\,\mathrm{m}$).

\section{PROPOSED METHOD}
\subsection{Overview}
\begin{figure*}[t]
    \centering
    \includegraphics[width=1.0\linewidth]{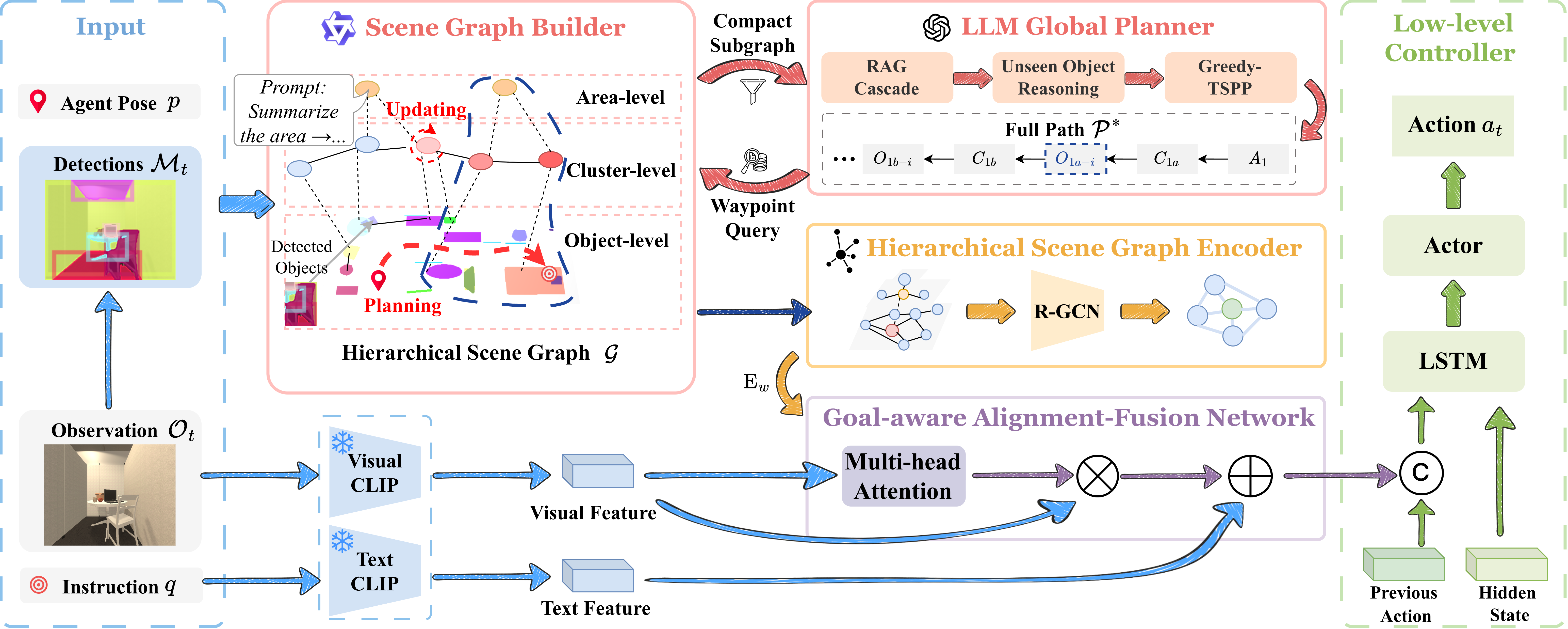}
    \caption{
        \textbf{Pipeline Overview}:
        (i) LLM-Guided Hierarchical Global Planner (H-GP) generates semantic waypoint sequences; (ii) Hierarchical Scene Graph Encoder (HSGE) grounds the plan in structured spatial-semantic representations; (iii) Goal-aware Alignment-Fusion Network (GAFN) aligns visual observations with graph-guided priors for efficient RL-based navigation.}
    \label{fig:overview}
    \vspace{-1.4em}
\end{figure*}

The proposed framework (Fig.~\ref{fig:overview}) explicitly bridges macroscopic semantic reasoning with reactive local control. The navigation pipeline operates as a continuous structural-to-perceptual cascade. Specifically, the abstract waypoint sequence generated by the LLM-driven \textbf{H}ierarchical \textbf{G}lobal \textbf{P}lanner (\textbf{H-GP}) is structurally encoded into topology-aware embeddings by the \textbf{H}ierarchical \textbf{S}cene \textbf{G}raph \textbf{E}ncoder (\textbf{HSGE}). These structural priors are then adaptively fused with real-time egocentric observations via the \textbf{G}oal-\textbf{A}ware \textbf{A}lignment-\textbf{F}usion \textbf{N}etwork (\textbf{GAFN}). Finally, the unified state representation is fed into an attentive Actor-Critic~\cite{A3C} network featuring a two-layer LSTM~\cite{LSTM} to maintain temporal coherence and generate the action policy $\pi_\theta(a_t \mid h_t,q)$.

\subsection{LLM-Guided Global Planning over Hierarchical Scene Graphs}
Inspired by recent advances that extend RAG to embodied environments~\cite{NavRAG,EmbodiedRAG}, we develop an LLM-driven global planner that operates directly over hierarchical scene graphs to enable semantically grounded long-horizon planning. By distilling high-level semantic priors into a structured navigation blueprint, this planner significantly reduces redundant exploration and provides a robust, interpretable guidance signal for downstream execution in unfamiliar environments.

\subsubsection{\textbf{Hierarchical Scene Graph Builder}}

We first establish a hierarchical scene graph $\mathcal{G} = (\mathcal{V}, \mathcal{E})$ with object-level nodes $v_i \in \mathcal{V}_{\text{object}}$ forming the foundational level~\cite{conceptgraphs}. Nodes follow a unified tuple $v_i = (L_i, p_i, f_i, l_i, \mathcal{D}_i)$, representing the semantic category, 3D centroid, visual embedding, abstraction level, and a VLM-generated summary, respectively. Crucially, $\mathcal{D}_i$ is set to empty ($\emptyset$) for object nodes.

To maintain real-time efficiency during active exploration, the scene hierarchy is incrementally constructed. Newly detected nodes either initialize high-level clusters or merge into existing ones based on a hybrid similarity metric $S_{ij}$, evaluated against nodes within the same abstraction level:
\begin{equation}
S_{ij} = (1 - \alpha) S^\text{spatial}_{ij} + \alpha S^{\text{semantic}}_{ij} \,,
\end{equation}
where $S^\text{spatial}_{ij}$ denotes the spatial similarity derived via a Gaussian kernel over the Euclidean distance, $S^{\text{semantic}}_{ij}$ represents the cosine similarity between visual embeddings, and $\alpha$ acts as the balancing weight. 

Upon merging, the cluster's spatial and visual attributes ($p_k, f_k$) are updated via online averaging of its children. To strictly bound computational overhead, the summary $\mathcal{D}_k$ is refreshed solely upon significant semantic drift.

Heterogeneous edges $\mathcal{E}$ capture the environment's topology, comprising distance-constrained intra-level spatial links and bidirectional inter-level hierarchical edges (semantic inclusion). This multi-relational connectivity explicitly encodes relation types and metric distances, providing a structural foundation for downstream R-GCN encoding (Sec.~\ref{sec:rgcn_encoder}) and LLM-driven planning (Sec.~\ref{sec:global_planning}).

\subsubsection{\textbf{Retrieval-Augmented Compact Subgraph}}
\label{sec:compact_subgraph}
During navigation, the hierarchical scene graph $\mathcal{G}$ undergoes continuous, high-frequency topological updates. To prevent policy oscillation and circumvent the severe latency of querying the LLM at every step, we explicitly decouple global reasoning from reactive control. The global planner operates asynchronously, caching waypoint sequences and triggering re-planning solely upon subgoal attainment or deadlocks.

To filter task-irrelevant context, we extract an instruction-relevant subgraph $\mathcal{G}_Q$.
For each candidate node $v$, we define a context-augmented semantic prior $s_v$ by combining its node-instruction semantic similarity with a hierarchical gain from its ancestor nodes.
To favor spatially coherent and topologically informative candidates, we further aggregate the semantic priors of its spatial neighborhood $\mathcal{N}(v)$:
\begin{equation}
 \mathcal{S}(v) =
 \Big(1+\beta \cdot \frac{1}{|\mathcal{N}(v)|}
 \sum_{u\in\mathcal{N}(v)} s_u \Big)s_v ,
\end{equation}
where $\beta$ controls the strength of neighborhood-based spatial boosting.
The target-relevant subgraph $\mathcal{G}_Q$ is induced by the top-$K$-ranked nodes, providing a compact and spatially coherent context for LLM-guided semantic planning.

\subsubsection{\textbf{LLM-driven Global Planning}}
\label{sec:global_planning}

Operating over $\mathcal{G}_Q$, the LLM acts as a semantic reasoning engine, infusing the hierarchical structure with rich commonsense priors by synthesizing functional summaries for abstracted areas and conducting hypothetical reasoning for unseen targets. 

Driven by these semantic foundations, we formulate the global planning task over $\mathcal{G}_Q$ as a Traveling Salesperson Path Problem ($\text{TSPP}$). To sequence these targets effectively, we propose a hierarchical coherence weight $W(v_i, v_j)$ that balances semantic desirability against geometric cost:
\begin{equation}
W(v_i, v_j) = \lambda_{\text{sem}} \cdot S_{\text{multi-level}}(v_i, v_j) - \lambda_{\text{dist}} \cdot D_{\text{geo}}(v_i, v_j) \,.
\end{equation}
$S_{\text{multi-level}}(v_i, v_j)$ denotes weighted sum of multi-level similarities. Leveraging aforementioned LLM summaries, it captures functional flow across abstract layers (e.g., a transition from $\texttt{Stove}$ to $\texttt{Sink}$ yields high coherence due to the shared $\texttt{Kitchen}$ context).
$D_{\text{geo}}$ penalizes excessive travel distance, governed by the balancing factors $\lambda_{\text{sem}}$ and $\lambda_{\text{dist}}$. 

The optimal sequence $\mathcal{P}^{*}$ is obtained via a Greedy-TSPP heuristic that iteratively selects the next unvisited node $v_{\text{next}}$ to maximize local coherence from the current node $v_c$:
\begin{equation}
v_{\text{next}} = \underset{v_i \in \mathcal{V}_{\text{cand}}}{\arg\max}\, W(v_c, v_i) \,,
\end{equation}
where $\mathcal{V}_{\text{cand}}$ represents the set of unvisited high-level candidate nodes. Ultimately, the regional trajectory $\mathcal{P}^*$ is sequentially grounded into object-level waypoints via top-down traversal of the hierarchical edges. Crucially, the HSGE instantiates each waypoint as a structure-aware embedding $E_w$, directly translating the LLM abstract reasoning into dense multi-modal guidance for the reactive policy.

\subsubsection{\textbf{Unseen Object Reasoning}}
For targets absent from the current view, SAGE-Nav leverages the LLM's broad internal knowledge to perform hypothetical probabilistic reasoning, inferring the most likely semantic cluster for a long-tail object (e.g., searching for a $\texttt{Remote}$ near a $\texttt{Sofa}$). This process enables target-guided exploration based on semantic co-occurrence. 

For an unseen or not-yet-observed target $q$, we introduce an LLM-induced semantic compatibility prior $\phi(q,c)$ over candidate clusters $c\in\mathcal{V}_{\text{cluster}}$:
\begin{equation}
 c^\ast =
 \underset{c \in \mathcal{V}_{\text{cluster}}}{\operatorname{argmax}}
 \ \phi(q,c),
\end{equation}
where $\phi(q,c)$ reflects the contextual co-occurrence likelihood between the target category $q$ and the cluster-level semantic summary $\mathcal{D}(c)$.
The most plausible cluster $c^\ast$ instantiates a virtual object node $v_{\text{virtual}}$ at the corresponding centroid $p(c^\ast)$, which is dynamically integrated into the Greedy-TSPP optimization process, encouraging the agent to prioritize semantically relevant search regions.

\subsection{Hierarchical Scene Graph Encoder}
\label{sec:rgcn_encoder}
\begin{figure}[t]
    \centering
    \includegraphics[width=1.0\columnwidth]{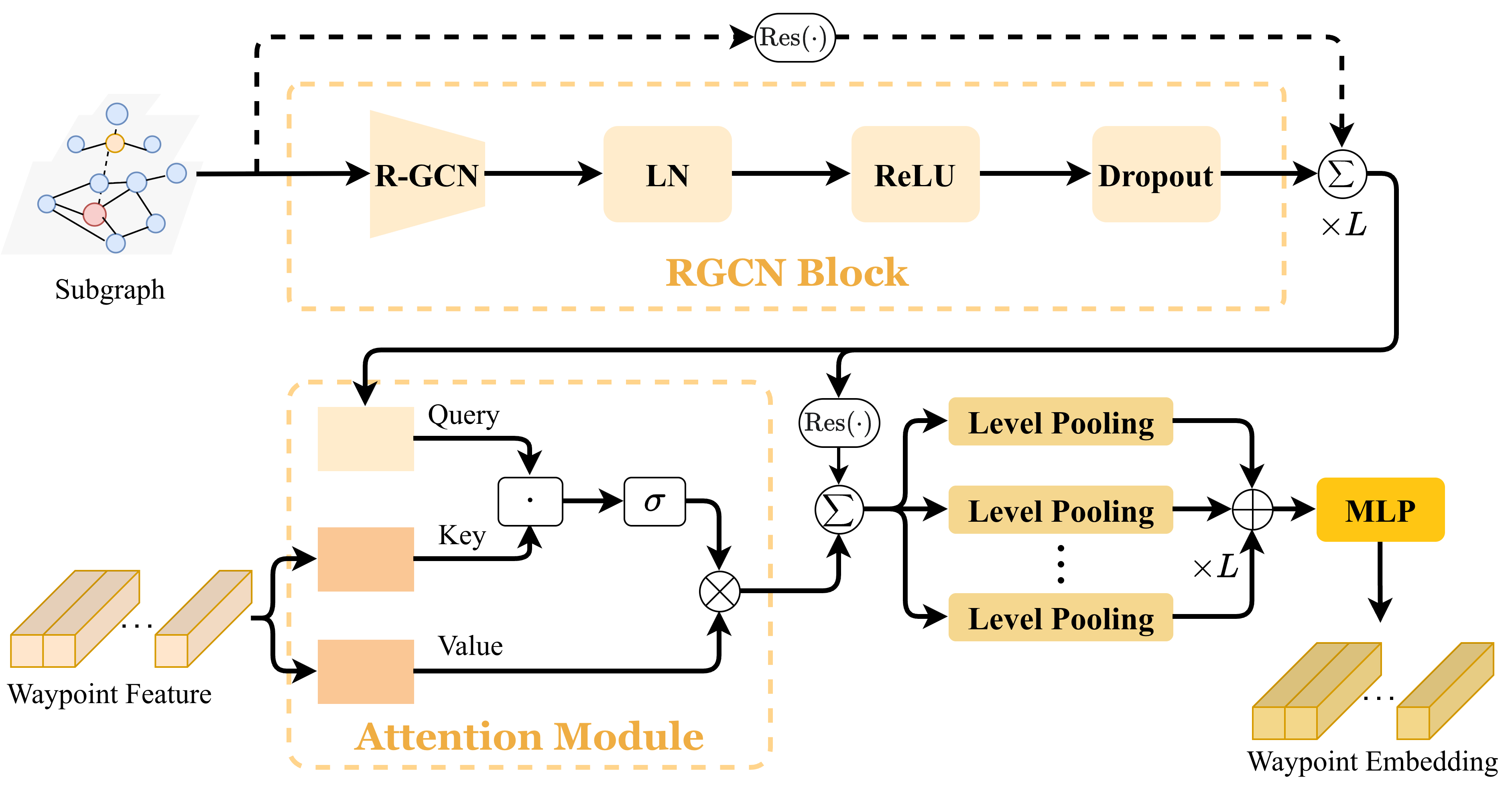}
    \caption{\textbf{HSGE Architecture.} A multi-layer R-GCN encodes structural relations, a residual attention module for task alignment and level-wise pooling produces the final structure-aware waypoint embedding $\mathbf{E}_w$.}
    \label{fig:hsge}
    \vspace{-1.5em}
\end{figure}

The HSGE functions as a structural cognitive map, designed to mitigate the inherent partial observability of embodied agents during navigation. As illustrated in Fig.~\ref{fig:hsge}, rather than treating the scene as a collection of isolated detections, the HSGE encodes the hierarchical scene graph $\mathcal{G}$ into persistent topological embeddings that maintain the scene's spatial-semantic context even when objects are outside the agent's immediate line-of-sight.

To establish a foundation for this structural memory, the feature of node $v_i \in \mathcal{V}$ is instantiated as $\mathbf{x}_i=[\mathbf{f}_i; \psi(\mathbf{p}_i)]$, where $\mathbf{f}_i$ and $\psi(\mathbf{p}_i)$ denote the CLIP~\cite{CLIP} semantic embedding and a sinusoidally encoded 3D coordinates, respectively.

\subsubsection{\textbf{Intra-level Relational Encoding}} 

To capture the heterogeneous topology within each abstraction level $\ell$, we employ a specialized Relational Graph Convolutional Network (R-GCN)~\cite{RGCN}. By abstracting the standard message-passing mechanism as a parameterized relational operator, we formulate the structure-aware embedding for node $v$ as:
\begin{equation}
    h_{v}^{(\ell)} = \mathcal{F}_{\text{R-GCN}}^{(\ell)} \big( \mathbf{x}_v, \mathcal{E}^{(\ell)} \big) + W_{\text{res}} \mathbf{x}_v \,,
\label{eq:RGCN_encoder}
\end{equation}
where $\mathcal{F}_{\text{R-GCN}}^{(\ell)}$ denotes the sequential application of the core R-GCN convolution, Layer Normalization (LN), ReLU activation, and Dropout (Fig.~\ref{fig:hsge}). $\mathcal{E}^{(\ell)}$ denotes the intra-level heterogeneous edge set. Crucially, the residual term $W_{\text{res}} \mathbf{x}_v$ projects the initial semantic anchors into the encoded space. This design effectively mitigates over-smoothing and preserves the raw multi-modal context during deep graph propagation.

Accordingly, given the active waypoint designated by the global planner (Sec.~\ref{sec:global_planning}) within the extracted subgraph (Sec.~\ref{sec:compact_subgraph}), we seamlessly derive its corresponding intrinsic embedding $\mathbf{x}_w$ via Eq.~\eqref{eq:RGCN_encoder}.

\subsubsection{\textbf{Waypoint-Conditioned Infusion and Encoding}} 
To ensure that the topological encoding is guided by the immediate navigation step, the R-GCN outputs are explicitly conditioned on the active waypoint in sequence $\mathcal{P}^{*}$. 
We define an alignment weight $\alpha_v \propto \exp(\mathbf{x}_w^{\top} {h_v^{(\ell)}})$ to represent the structural relevance of each node $v$ to the active waypoint. The node representations are subsequently refined through a residual attention mechanism:

\begin{equation}
\tilde{h}_v^{(\ell)} = h_v^{(\ell)} + \alpha_v \mathbf{x}_w\,.
\end{equation}

Following this waypoint-aware infusion, we aggregate the augmented embeddings across the subgraph to formulate a unified representation. By concatenating the average-pooled features from all abstraction levels $\ell \in \{0, \dots, L\}$, we compute the waypoint embedding $\mathbf{E}_w$ through an MLP:
\begin{equation}
\mathbf{E}_w = \text{MLP} \left( \Big[ \text{AvgPool} \big( \{ \tilde{h}_v^{(\ell)} \} \big) \Big]_{\ell=0}^L \right) \,,
\end{equation}
where $[\cdot]_{\ell=0}^L$ denotes the channel-wise concatenation across hierarchical levels. 

By seamlessly fusing multi-scale spatial relations with task-driven waypoint semantics, this localized embedding $\mathbf{E}_w$ inherently captures both the target's intrinsic properties and its rich topological neighborhood, providing a deterministic and structure-aware signal for the low-level policy.

\subsection{Goal-Aware Alignment-Fusion Network}

To bridge global semantic planning and reactive local control, GAFN performs multi-modal manifold alignment between real-time egocentric perception $\mathbf{F}_v$ and structure-aware waypoint embedding $\mathbf{E}_w$ from the HSGE.

\subsubsection{\textbf{Structure-Infused Cross-Attention}}
Instead of a rigid feature concatenation, we first ground the abstract topological guidance within the perceptual space. Using cross-attention, the visual feature $\mathbf{F}_v$ serves as the query, while the waypoint embedding $\mathbf{E}_w$ serves as the key and the value. We then compute a structure-infused representation $\mathbf{F}_a = \text{CrossAttn}(\mathbf{F}_v, \mathbf{E}_w)$~\cite{attention}. This ensures the latent visual features are aligned with the geometric and semantic constraints of the target region.

\subsubsection{\textbf{Alignment-Aware Gated Fusion}}

To balance reactive obstacle avoidance (via $\mathbf{F}_v$) and goal-directed navigation (via $\mathbf{F}_a$), we compute a real-time semantic alignment score $\alpha_t = \cos(\mathbf{F}_v, \mathbf{E}_w)$. To encourage a behavioral shift as the agent approaches the target, we formulate an adaptive gating vector $\lambda$ designed to structurally modulate the fusion process with an explicit inductive bias. Specifically, an MLP-learned base gate is weighted by a penalty term governed by $\alpha_t$:
\begin{align}
\lambda(\alpha_t) &= \sigma \big( \text{MLP}([\mathbf{F}_v; \mathbf{F}_a; \mathbf{E}_w; \alpha_t]) \big) \cdot \big(1 - \eta \sigma(\kappa \alpha_t)\big) \,, \\
\mathbf{F}_{f} &= \lambda(\alpha_t) \odot \mathbf{F}_v + \big(1 - \lambda(\alpha_t)\big) \odot \mathbf{F}_a \,,
\end{align}
where $\sigma$ is the Sigmoid activation, $\odot$ denotes element-wise multiplication, and $\eta, \kappa$ are scaling hyperparameters (empirically set to 0.5 and 2.0) designed to smoothly modulate the alignment weighting. Empirically, as the agent's visual perception progressively aligns with the designated waypoint (increasing $\alpha_t$), the explicit penalty term $(1 - \eta \sigma(\kappa \alpha_t))$ drives the overall gating $\lambda$ to decrease. This structural design incentivizes the network to shift its reliance from raw reactive perception to the rich structural context $\mathbf{F}_a$, stabilizing the final approach before feeding $\mathbf{F}_{f}$ into the downstream policy.

\subsection{Reinforcement Learning Optimization}
\begin{figure}[t]
    \centering
    \includegraphics[width=1.0\columnwidth]{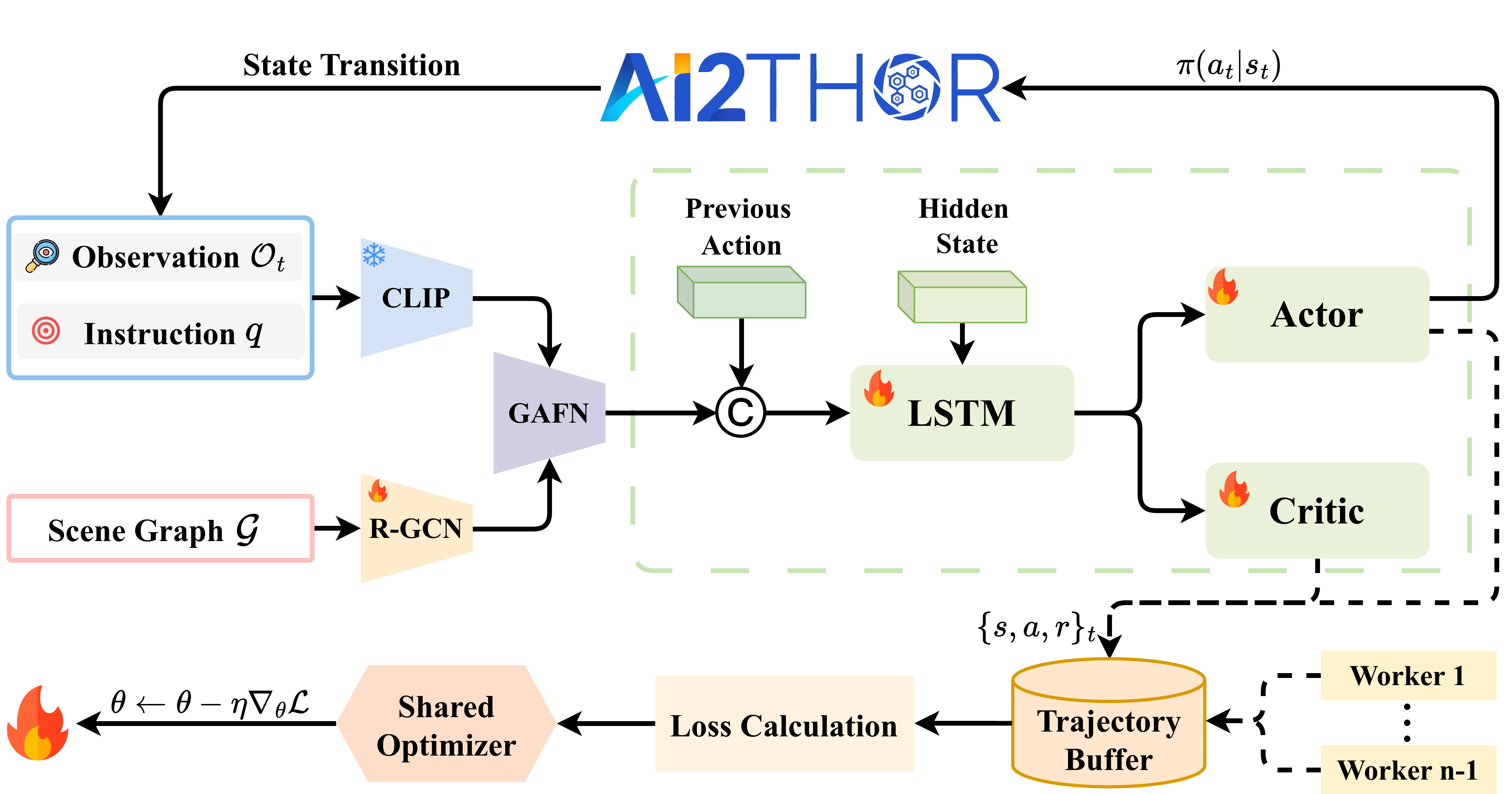}
    \caption{
    \textbf{The asynchronous A3C training architecture.} Parallel workers collect trajectories to compute losses. A shared optimizer backpropagates gradients—visually denoted by the fire symbol—to update network, whereas frozen modules are marked by the snowflake symbol.}
\label{fig:train_utils}
    \vspace{-1.5em}
\end{figure}

As illustrated in Fig.~\ref{fig:train_utils}, the fused representation from GAFN is flattened and fed into a standard two-layer LSTM-based~\cite{LSTM} Actor-Critic network. This recurrent module maintains temporal coherence, generating the action distribution $\pi_\theta(a_t|s_t)$ and the state value estimation $V_\phi(s_t)$.

The navigation reward $R_{\text{nav}}$ is formulated as a composite signal to simultaneously encourage goal-reaching, exploration, and trajectory efficiency:
\begin{equation}
    R_{\text{nav}} = 5 \cdot \mathbb{I}_{vis} + 0.01 \cdot \mathbb{I}_{fwd} + \max \big( \lambda \Delta d_t, 0 \big) - 0.01 \cdot (1 - \mathbb{I}_{vis}) \,,
\end{equation}
where $\mathbb{I}_{vis}$ and $\mathbb{I}_{fwd}$ indicate target visibility and forward movement ($a_t = \text{MoveAhead}$), respectively. $\Delta d_t = d_{t-1} - d_t$ denotes the reduction in Euclidean distance to the target at step $t$, and the final term imposes a constant step penalty for target invisibility and redundant actions.

Driven by the collected trajectories, the framework is optimized using the standard A3C objective~\cite{A3C}, combining policy gradient, value, and entropy regularization losses.

\section{EXPERIMENTS}

\subsection{Experimental Setup}
\subsubsection{\textbf{Datasets}}
We evaluate the proposed framework across two widely used embodied simulation datasets: \textit{iTHOR}~\cite{AI2Thor} and \textit{RoboTHOR}~\cite{Robothor}. 
\textit{iTHOR} comprises 120 photorealistic indoor scenes evenly distributed among four room categories (\textit{kitchen, living room, bedroom, bathroom}). Each room exhibits unique furniture layouts and object distributions. We use the first 20 scenes of each category for training, the next 5 for validation, and the remaining 5 for testing. 
We consider 22 object categories as targets, ensuring at least four objects per room type.
\textit{RoboTHOR} contains 89 apartments—75 for training and validation and 14 for testing—with approximately 2.4$\times$ larger area and 5.5$\times$ longer trajectories than \textit{iTHOR}. Following prior study~\cite{ContextAwareGI}, we use 60 apartments for training, 5 for validation, and 10 for testing. 

\subsubsection{\textbf{Evaluation Metrics}}
To comprehensively assess navigation performance, we adopt three standard Object-Goal Navigation metrics~\cite{batra2020objectnav}: Success Rate (SR), Success weighted by Path Length (SPL), and Distance to Success (DTS). An episode is considered successful if the agent issues the \texttt{Done} action within a success threshold of $d_s = 1.5$m from the target. We report these metrics across all trajectories (\textit{ALL}) and challenging long-horizon scenarios with optimal path lengths of at least 5 meters ($L \ge 5$).

\subsubsection{\textbf{Implementation Details}}

We use frozen CLIP ViT-B/32~\cite{CLIP} for 512-D visual feature extraction and Qwen2.5-VL-7B~\cite{Qwen25VLTR} as the global reasoning backend. The HSGE employs a 3-layer R-GCN~\cite{RGCN} to process 544-D inputs (visual features concatenated with 32-D sinusoidal positional encodings) into 64-D waypoint embeddings. GAFN linearly projects flattened visual maps and waypoint embeddings into a 128-D latent space for alignment fusion, outputting a 64-D representation for the LSTM policy. Following CogNav~\cite{CogNav}, our online 3D scene graph integrates OpenSEED~\cite{OpenSEED} segmentation, multi-frame fusion, and VLM-driven relation reasoning, dynamically aggregating nodes via a $2.0$m distance and $\alpha=0.6$ semantic affinity threshold. 

The A3C~\cite{A3C} policy is trained for 6 million episodes using Adam (learning rate $1\times10^{-4}$). To accelerate the reinforcement learning, the hierarchical scene graphs for the training environments are pre-constructed offline. Consequently, the optimization exclusively focuses on tuning the learnable components, namely the Hierarchical Scene Graph Encoder (HSGE) and the LSTM-based policy network. RL objectives are balanced with a value weight $\lambda_v=0.5$, entropy weight $\lambda_e=0.01$, and discount factor $\gamma=0.99$. The progress reward multiplier is $\lambda=0.1$. Final evaluations report the best validation checkpoint across 1,000 test episodes. Notably, all experiments run on a single NVIDIA RTX 3090 GPU, ensuring the high-frequency reactive control loop remains strictly decoupled from asynchronous LLM planning overhead.

\begin{table*}[t]
    \centering
    \caption{COMPARISONS WITH BASELINE AND SOTA METHODS IN THE iTHOR AND ROBOTHOR}
    \label{table:full_comparison}
    \setlength{\tabcolsep}{3pt}
    \begin{NiceTabular*}{\textwidth}{@{\extracolsep{\fill}}lcccccccccccc}
        \CodeBefore
            \rowcolor[HTML]{F0F8FF}{17}
        \Body
        \toprule
        \multirow{3}{*}{Method} & \multicolumn{6}{c}{iTHOR} & \multicolumn{6}{c}{RoboTHOR} \\
        \cmidrule(lr){2-7} \cmidrule(lr){8-13}
        & \multicolumn{3}{c}{ALL} & \multicolumn{3}{c}{$L \ge 5$} & \multicolumn{3}{c}{ALL} & \multicolumn{3}{c}{$L \ge 5$} \\
        \cmidrule(lr){2-4} \cmidrule(lr){5-7} \cmidrule(lr){8-10} \cmidrule(lr){11-13}
        & SR$\uparrow$(\%) & SPL$\uparrow$(\%) & DTS$\downarrow(m)$ & SR$\uparrow$(\%) & SPL$\uparrow$(\%) &DTS$\downarrow(m)$ & SR$\uparrow$(\%) & SPL$\uparrow$(\%) &DTS$\downarrow(m)$ & SR$\uparrow$(\%) & SPL$\uparrow$(\%) &DTS$\downarrow(m)$ \\
        \midrule
        Random & 3.72 & 1.93 & 1.45 & 0.21 & 0.08 & 1.95 & 0.00 & 0.00 & 2.50 & 0.00 & 0.00 & 3.10 \\
        SAVN~\cite{SAVN} & 40.21 & 16.03 & 1.15 & 29.86 & 17.81 & 1.58 & 28.42 & 17.82 & 1.39 & 22.13 & 15.34 & 1.61 \\
        EOTP~\cite{EOTP} & 46.20 & 17.88 & 0.82 & 32.63 & 15.94 & 0.97 & 13.57 & 5.28 & 1.55 & 5.14 & 2.85 & 1.82 \\
        ORG~\cite{ORG} & 68.10 & 41.14 & 0.55 & 56.76 & 38.16 & 0.77 & 29.61& 19.23& 1.35 & 22.53 & 15.73 & 1.48 \\
        HOZ~\cite{HOZ} & 73.32 & 39.07 & 0.44 & 65.50 & 38.85 & 0.60 & 32.27& 20.48& 1.28 & 24.83 & 16.89 & 1.42 \\
        VTNet~\cite{VTNet} & 72.20 & 44.90 & 0.52 & 63.40 & 44.00 & 0.69 & 31.62 & 19.63& 1.30 & 23.48 & 16.02& 1.45 \\
        L-sTDE~\cite{L-sTDE} & 74.19 & 40.30 & 0.43 & 64.01 & 39.97 & 0.60 & 42.13 & 24.54& 1.05 & 32.04 & 17.44 & 1.20 \\
        DOA~\cite{DOA} & 76.61 & 44.86 & 0.48 & 68.22 & 44.71 & 0.62 & --- & --- & --- & --- & --- & --- \\
        TDANet~\cite{TDANet} & 78.20 & 30.60 & --- & 67.00 & 33.40 & --- & --- & --- & --- & --- & --- & --- \\
        AKGVP~\cite{AKGVP} & 73.63 & 40.66 & 0.44 & 63.51 & 39.63 & 0.60 & 39.69 & 25.84& 1.10 & 28.55 & 18.79 & 1.25 \\
        AKGVP-CI~\cite{AKGVP} & 76.78 & 39.63 & 0.35 & 65.45 & 39.01 & 0.53 & 44.53 & 27.61& 1.00 & 32.68 & 20.55 & 1.21 \\
        CGI-GAIL~\cite{ContextAwareGI} & 77.59 & \textbf{46.25} & --- & 69.18 & \textbf{46.10} & --- & 48.30 & 28.77 & --- & 36.69 & 21.89 & --- \\
        TSOG~\cite{TemporalSG} & 80.04 & 41.44 & 0.40 & 73.46 & 43.66 & 0.56 & 49.89 & 28.41 & 0.82 & 38.61 & 21.15 & 1.18 \\
        \midrule
        \textbf{SAGE-Nav (Ours)} & \textbf{82.47} & 42.34 & \textbf{0.32} & \textbf{77.22} & 43.73 & \textbf{0.43} & \textbf{52.35} & \textbf{30.12} & \textbf{0.72} & \textbf{40.82} & \textbf{22.95} & \textbf{1.10}\\
        \bottomrule
    \end{NiceTabular*}
    \vspace{0.5em}
    \raggedright \footnotesize \textit{Note:} `---' denotes metrics that are neither reported in the original publications nor reproducible for evaluation.
    \vspace{-1.5em}
\end{table*}

\subsection{Quantitative Results}

\textbf{General Navigation Performance.} 
Table~\ref{table:full_comparison} summarizes the quantitative results, with baseline metrics adopted from their original publications for fair comparison. In iTHOR, SAGE-Nav achieves state-of-the-art Success Rates (SR) of 82.47\% overall and 77.22\% in challenging long-horizon scenarios ($L \ge 5$), outperforming TSOG and CGI-GAIL by absolute margins of 3.76 and 8.04 percentage points, respectively. Although CGI-GAIL retains a marginally higher SPL due to its strict reliance on imitation learning from expert demonstrations, SAGE-Nav achieves highly competitive efficiency (43.73\%) purely through LLM-derived reasoning and topological guidance. Furthermore, in the structurally complex RoboTHOR environment, SAGE-Nav establishes a new state-of-the-art across all metrics, notably reaching 40.82\% SR and 22.95\% SPL in long-horizon tasks. This performance comprehensively  validates the robustness of our hierarchical priors and dynamic scheduling mechanism.

\begin{table}[t]
    \centering
    \caption{ZERO-SHOT PERFORMANCE COMPARISON AND ABLATION STUDY}
    \label{table:zeroshot}
    \renewcommand{\arraystretch}{1.15}
    \setlength{\tabcolsep}{9pt}
    \begin{NiceTabular}{lccc}
        \CodeBefore
            \rowcolor[HTML]{F0F8FF}{9}
        \Body
        \toprule
        \textbf{Navigator} & \textbf{SR$\uparrow$(\%)} & \textbf{SPL$\uparrow$(\%)} & \textbf{DTS$\downarrow$(m)} \\
        \midrule
        TargetDriven~\cite{targetdriven2017} & 19.34 & 8.81 & 1.20 \\
        HOZ~\cite{HOZ}                       & 47.54 & 15.05 & 0.76 \\
        L-sTDE~\cite{L-sTDE}                 & 54.75 & 16.64 & 0.62 \\
        AKGVP-CI~\cite{AKGVP}                   & 69.51 & 28.86 & 0.41 \\
        \midrule
        \textbf{w/o Unseen Inference}        & 73.84 & 33.50 & 0.40 \\
        \textbf{w/o LLM Reasoning}           & 73.30 & 33.19 & 0.42 \\
        \textbf{w/o Alignment Fusion}        & 70.65 & 30.80 & 0.47 \\
        \midrule
        \textbf{SAGE-Nav (Ours)}             & \textbf{75.05} & \textbf{34.05} & \textbf{0.38} \\
        \bottomrule
    \end{NiceTabular}
    \vspace{-1.0em}
\end{table}

\textbf{Zero-shot Navigation Performance.} We conduct zero-shot experiments to evaluate the system's ability to generalize to unseen object categories and infer their likely semantic contexts. Following the experimental protocols of prior studies~\cite{AKGVP}, we select six object categories :\texttt{Bowl}, \texttt{DeskLamp}, \texttt{Laptop},
\texttt{LightSwitch}, \texttt{Plate}, and \texttt{StoveBurner} that are excluded during the training phase but present within the test set. As shown in Table~\ref{table:zeroshot}, SAGE-Nav outperforms all baseline methods across all metrics, achieving a 75.05\% SR, 34.05\% SPL and $0.38\,\mathrm{m}$ DTS. This substantial improvement over the previous SOTA validates that our hierarchical representation effectively captures transferable semantic priors for unfamiliar environments.

\textbf{Ablation Study on Zero-shot Generalization.} Component analysis further elucidates the hierarchical contributions of each module to the zero-shot capability. The most substantial degradation occurs upon removing the Alignment Fusion module (SR drops to 70.65\%), underscoring that fine-grained synchronization between semantic waypoints and egocentric vision is fundamental for cross-domain adaptation. Replacing the LLM with heuristic rules that select waypoints based solely on graph scores highlights the critical role of common-sense priors, which rigid graph searches fail to capture. Finally, the slight performance drop observed in w/o Unseen Inference validates the utility of hypothetical probabilistic reasoning in refining the localization of long-tail targets.

\begin{table}[t]
    \centering
    \caption{PERFORMANCE VS. EFFICIENCY TRADE-OFF IN ROBOTHOR}
    \label{table:robothor_efficiency}
    \renewcommand{\arraystretch}{1.15}
    \setlength{\tabcolsep}{4.5pt}
    \begin{NiceTabular}{lcccc}
        \CodeBefore
            \rowcolor[HTML]{F0F8FF}{4}
        \Body
        \toprule
        \textbf{Method} & \textbf{SR$\uparrow$(\%)} & \textbf{SPL$\uparrow$(\%)} & \textbf{Latency$\downarrow$(s)} & \textbf{LLM Calls$\downarrow$} \\
        \midrule
        CogNav~\cite{CogNav}  & \textbf{54.6} & 24.3 & 2.20 & 15 \\
        SG-Nav~\cite{SG-Nav}  & 47.5 & 24.0 & 0.85 & 32 \\
        \midrule
        \textbf{SAGE-Nav (Ours)}  & 52.4 & \textbf{30.1} & \textbf{0.42} & \textbf{9.1} \\
        \bottomrule
    \end{NiceTabular}
    \vspace{-1.0em}
\end{table}

\textbf{Performance-Efficiency Analysis.}
Table~\ref{table:robothor_efficiency} evaluates the performance-efficiency trade-off in RoboTHOR using two metrics: \textit{Latency}, the average wall-clock time per execution step, and \textit{LLM Calls}, the average number of LLM invocations per episode.
While CogNav achieves the highest SR (54.6\%), its reliance on heavy visual and multimodal foundation models at each step leads to high action latency ($2.20\,\mathrm{s}$).
SG-Nav reduces this latency to $0.85\,\mathrm{s}$ but requires frequent LLM queries (32 calls per episode).
In contrast, SAGE-Nav invokes the LLM planner sparsely and asynchronously, requiring only 9.1 LLM calls per episode while maintaining an efficient reactive control loop.
It achieves the lowest latency ($0.42\,\mathrm{s}$), the highest SPL (30.1\%), and a competitive SR (52.4\%).
These results show that decoupling high-level semantic planning from local reactive control improves inference efficiency while preserving navigation performance.

\subsection{Qualitative Results }
\begin{figure}[t]
    \centering

    \includegraphics[width=1.0\columnwidth]{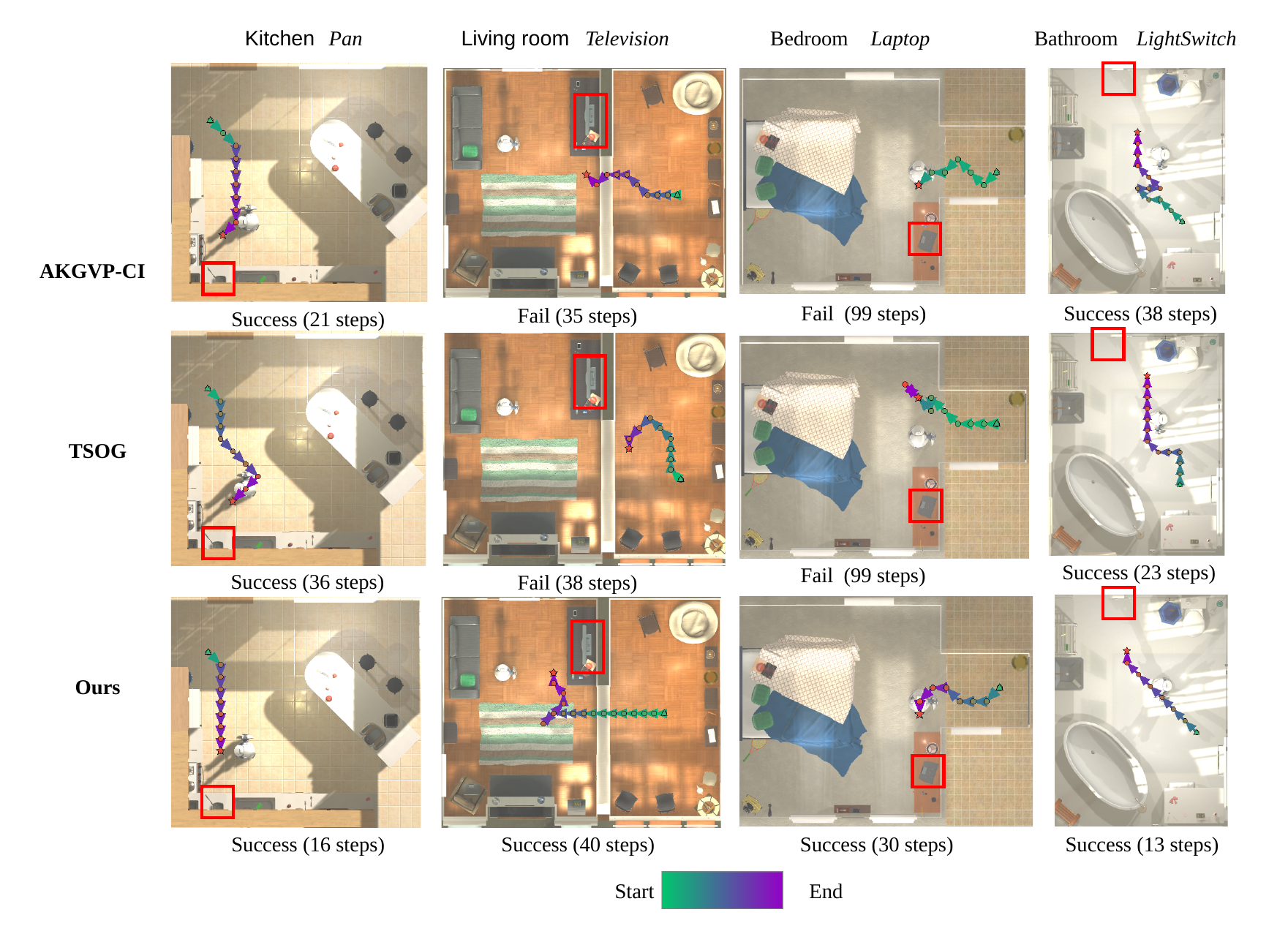}
    \caption{Visualization of the agent trajectories in unfamiliar scenes in the AI2-THOR environment.}
    \label{fig:trajectory_vis}
\end{figure}
Figure~\ref{fig:trajectory_vis} visualizes navigation trajectories in the simulator, comparing SAGE-Nav with AKGVP~\cite{AKGVP} and TSOG~\cite{TemporalSG} across four scenarios. The baseline algorithms exhibit suboptimal path planning, resulting in redundant paths and detours. Furthermore, they are prone to getting stuck in rotational loops or prematurely terminating episodes when targets are occluded. In contrast, SAGE-Nav demonstrates superior path efficiency. Crucially, leveraging waypoint guidance for hard-to-find targets, our method improves both the navigation success rate and overall robustness.
\subsection{Ablation Studies }   

\begin{table}[t]
    \caption{IMPACTS OF DIFFERENT MODULES}
    \label{Table:ablation study}
    \centering
    \setlength{\tabcolsep}{4pt}
    \resizebox{\linewidth}{!}{%
    \begin{NiceTabular}{l ccc ccc}
        \CodeBefore
            \rowcolor[HTML]{F0F8FF}{9}
        \Body
        \toprule
        \multirow{2}{*}{ID} & \multicolumn{3}{c}{Method} & \multicolumn{3}{c}{ALL / L $\ge$ 5} \\
        \cmidrule(r){2-4} \cmidrule(l){5-7}
        & H-GP & HSGE & GAFN & SR$\uparrow$(\%) & SPL$\uparrow$(\%) & DTS$\downarrow$ (m) \\
        \midrule
        1 & & & & 74.21 / 65.53 & 40.23 / 39.75 & 0.48 / 0.62 \\
        2 & \checkmark & & & 76.80 / 71.33 & 41.93 / 43.71 & 0.41 / 0.54 \\
        3 & & \checkmark & & 75.62 / 67.15 & 42.50 / 41.87 & 0.45 / 0.57 \\
        \midrule
        4 & \multicolumn{3}{l}{SAGE-Nav w/o H-GP} & 79.52 / 74.35 & 41.36 / 42.81 & \textbf{0.31} / 0.46 \\
        5 & \multicolumn{3}{l}{SAGE-Nav w/o HSGE} & 80.15 / 75.10 & 40.85 / 42.26 & 0.35 / 0.48 \\
        6 & \multicolumn{3}{l}{SAGE-Nav w/o GAFN} & 78.65 / 73.95 & \textbf{43.73 / 44.94} & 0.39 / 0.50 \\
        \midrule
        7 & \multicolumn{3}{l}{\textbf{SAGE-Nav (full)}} & \textbf{82.47 / 77.22} & 42.34 / 43.73 & 0.32 / \textbf{0.43} \\
        \bottomrule
    \end{NiceTabular}%
    }
    \vspace{-1.0em}
\end{table}

Table~\ref{Table:ablation study} summarizes the ablation study of proposed modules. The full SAGE-Nav (ID 7) achieves the best SR, particularly in long-horizon tasks ($L\ge5$).  %
Incremental integration (IDs 1-3) confirms that hierarchical planning and encoding significantly improve SR and DTS, emphasizing the synergy of structural reasoning. Specifically, the inclusion of H-GP (ID 2) provides crucial commonsense priors for target localization, improving SR by 2.59\% over the baseline.  %
The second part (IDs 4-6) validates the necessity of each component through separate removal. Eliminating the HSGE (ID 5) leads to a notable drop in success rates, confirming that multi-scale structural embeddings are vital for capturing scene-level global correlations. Notably, when GAFN is removed and substituted with simple feature concatenation (ID 6), the model exhibits a marginal SPL gain but a concurrent deterioration in SR and DTS. This suggests that while GAFN introduces a conservative exploration constraint to ensure alignment, it is essential for balancing high-level structural guidance with real-time perception, thereby enhancing navigational robustness in complex environments.

\subsection{Limitations}
\begin{figure}[t]
    \centering
    \includegraphics[width=1.0\columnwidth]{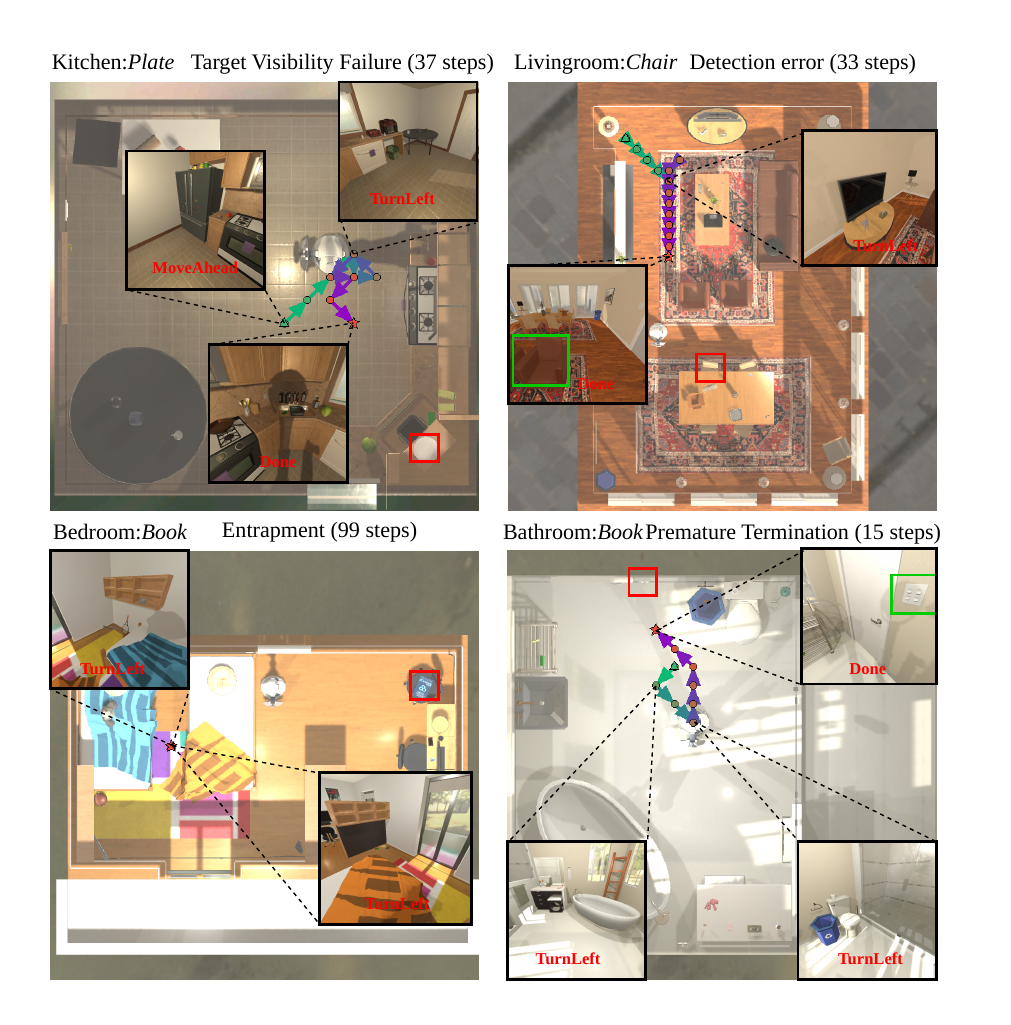}
    \caption{Failure case visualizations. The target object is highlighted with a red bounding box, while the agent-detected target is marked in green across egocentric observations at key navigation actions.}
    \label{fig:failure cases}
\vspace{-1.0em}
\end{figure}

We analyze the failure cases (Fig.~\ref{fig:failure cases}), which fall into four categories: (a) Target Visibility Failure, where the agent terminates despite the target (e.g., plates on high shelves) being outside the field of view; (b) Detection Error, caused by false positives in the detector between semantically similar objects like chairs and couches; (c) Entrapment, where the agent struggles to extricate itself from complex or narrow starting positions; and (d) Premature Termination, where the agent misjudges spatial proximity and stops before reaching the success threshold. To address these limitations, future work will explore integrating robust open-vocabulary detectors, developing active vision strategies for proactive viewpoint adjustment, and leveraging 3D volumetric representations to enhance obstacle avoidance in constrained spaces.

\section{CONCLUSIONS}
We propose SAGE-Nav, a hierarchical navigation framework that bridges the global reasoning prowess of LLMs with the structural expressiveness of scene graphs. By distilling LLM-derived semantic priors into a sequence of interpretable waypoints, our method effectively overcomes the long-horizon reasoning limitations inherent in traditional monolithic models. The precise alignment between real-time perception and structured guidance, facilitated by the HSGE and GAFN modules, ensures robust execution in diverse environments.
Quantitative evaluations in iTHOR and RoboTHOR demonstrate that SAGE-Nav achieves state-of-the-art performance, notably reaching an 82.47\% success rate and significantly enhancing zero-shot generalizability. Qualitative analysis further confirms its superior planning and obstacle avoidance capabilities in complex scenarios. Moving forward, we aim to augment the system's robustness by integrating active perception and generative navigation policies to further enhance extrication capabilities in constrained real-world settings.
\renewcommand*{\bibfont}{\footnotesize} 
\printbibliography[title={References}] 
\end{refsection}

\balance

{
\AtNextBibliography{\scriptsize}
\printbibliography
}

\end{document}